# Forecasting the successful execution of horizontal strategy in a diversified corporation via a DEMATEL-supported artificial neural network - A case study

Hossein Sabzian[1, *], Hossein Gharib[2], Javad Noori[3], Mohammad Ali Shafia [4], Mohammad Javad Sheikh[5]

[1] Department of Progress Engineering, Iran University of Science and Technology, Tehran, Iran

[2] Ministry of Health and Medical Education, Tehran, Iran

[3] Research Institute for Science, Technology and Industrial Policy, Sharif University of Technology, Tehran, Iran

[4] Department of Progress Engineering, Iran University of Science and Technology, Tehran, Iran

[5] Department of humanities, Shahed University, Tehran, Iran

*Corresponding author: hossein_sabzian@pgre.iust.ac.ir

## Abstract

Nowadays, competition is getting tougher as market shrinks because of financial crisis of the late 2000s. Organizations are tensely forced to leverage their core competencies to survive through attracting more customers and gaining more efficacious operations. In such a situation, diversified corporations which run multiple businesses have opportunities to get competitive advantage and differentiate themselves by executing horizontal strategy. Since this strategy completely engages a number of business units of a diversified corporation through resource sharing among them, any effort to implement it will fail if being not supported by enough information. However, for successful execution of horizontal strategy, managers should have reliable information concerning its success probability in advance. To provide such a precious information, a three-step framework has been developed. In the first step, major influencers on successful execution of horizontal strategy have been captured through literature study and interviewing subject matter experts. In the second step through the decision making trial and evaluation laboratory (DEMATEL) methodology, critical success factors (CSFs) have been extracted from major influencers and a success probability assessment index system (SPAIS) has been formed. In the third step, due to the statistical nature (multivariate and distribution free) of SPAIS, an artificial neural network has been designed for enabling organizational managers to forecast the success probability of horizontal strategy execution in a multi-business corporation far better than other classical models.

# 1. Introduction

In today's business world that competition has become harsher; firms have to learn ways of attaining sustained competitive advantage in order to survive. Therefore, understanding the sources of sustainable competitive advantage has become a main area of research for organizational strategists (Barney 1991; Ansoff 1957; David 2011; De Wit and Meyer 2010; Jacobides 2010). One way for gaining competitive advantage in large and diversified corporations is the application of horizontal strategy in order to share resources among business units (Porter 1985). Diversified corporations that run multiple businesses can use horizontal strategy to differentiate themselves and gain competitive advantage (Takaoka 2011). As a powerful way of synergy creation, horizontal strategy deals with resource sharing among those business units that are related in terms of skill and activity. Synergy creation through horizontal strategy enables diversified firms to gain significant competitive advantages and in turn increase profitability of the corporation as a whole(Shovareini, Alborzi, and Mohammad, n.d.).

By identifying and sharing related resources among business units of a corporation, (both tangible and intangible) horizontal strategy brings about a great deal of synergy and then results in significant competitive advantages for the entire corporation. Thus, this study is aimed at proposing a three-step framework by which managers of diversified corporations can get enabled to forecast how much a horizontal strategy can be successfully executed in a diversified corporation.

In the first step of the framework, major drivers influencing the success probability of horizontal strategy have been extracted through literature study and direct interviews with expert. In the second step, as a powerful MCDM method, DEMATEL has been used to identify the CSFs from all extracted influential drivers. The results of DEMATEL are used to form a success probability assessment index system (SPAIS). In the third step, an Iranian diversified corporation has been selected as a case study. According to SPAIS developed in the second step and participation of selected number of corporation's personnel, an artificial neural network (ANN) has been designed which can be used as a decision support system (DSS) to enable organizational managers to successfully execute the horizontal strategy.

## 2. Theoretical Background

### 2.1. Diversified Corporation and Corporate Strategy

A diversified corporation is an entity that runs multiple business units. Generally, according to Hiroyuki and Tada Kagno (2003) there are three main reasons for diversification of a business: economy of risk, dispersion of risk, and the economy of growth. Economy of scale is achieved when companies reduce costs through sharing resources among multiple business units. Having multiple business units in turn mitigates the risk of environmental change and this is what dispersion of risk is all about. Economy of growth is about those economic advantages that growth itself can cause (Takaoka 2011). The overall scope and direction of a corporation and the way in which its various business operations work together to achieve particular goals are dealt with by corporate strategy.

As a buzzword in business world, corporate strategy is fundamentally about two functions: I: selection of corporate business units and II: management of diverse business units and coordination of their strategies. For a long time, the first function has been the focus of attention of several strategists, but since recent world financial crisis, corporate strategists have shifted their attention towards the second function far more than before. This is largely because of the fact that attaining competitive advantage deeps heavily on the way corporations manage and coordinate their business units.

### 2.2. Types of diversified corporations in terms of organizational structure

Corporations can be classified into three types of A, B and C in terms of their structure. Corporation of type A is a single company entailing two or more business units as shown in figure 1:

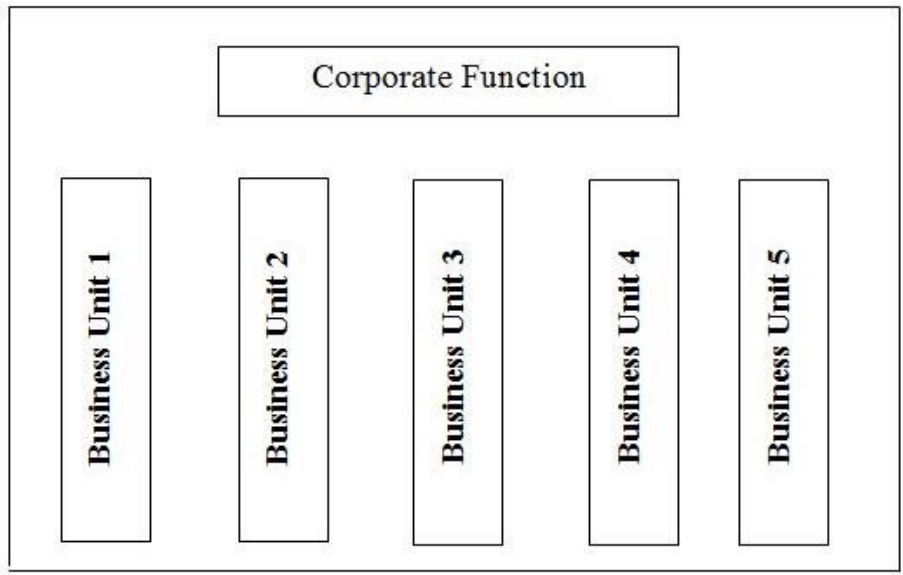

Figure 1: type A (Takaoka 2011)

Corporation of type B is a main company that runs a primary business and one or more subsidiaries that each runs other businesses as shown in figure 2:

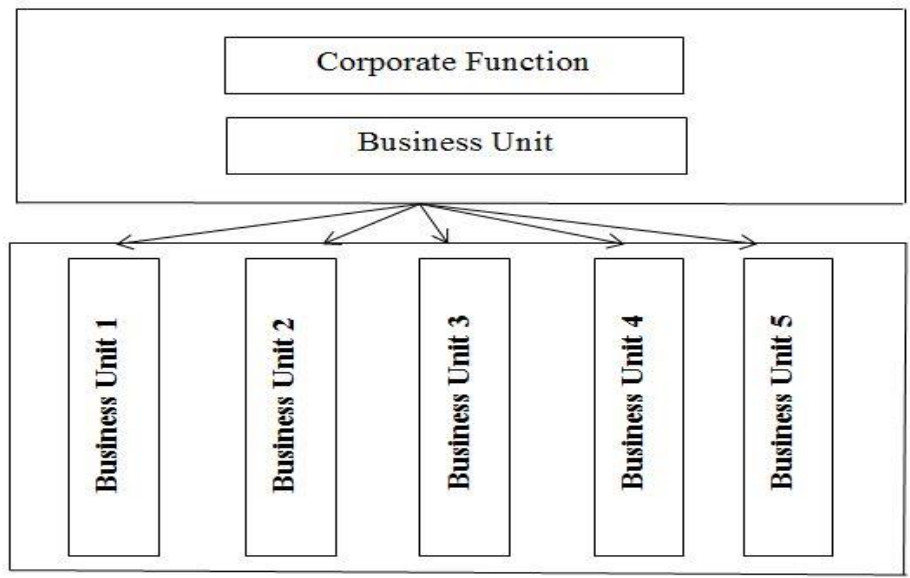

Figure 2: type B (Takaoka 2011)

Corporation of type C is one pure holding company that focuses on managing corporate strategy with two or more subsidiaries that run actual operational businesses as show in figure 3:

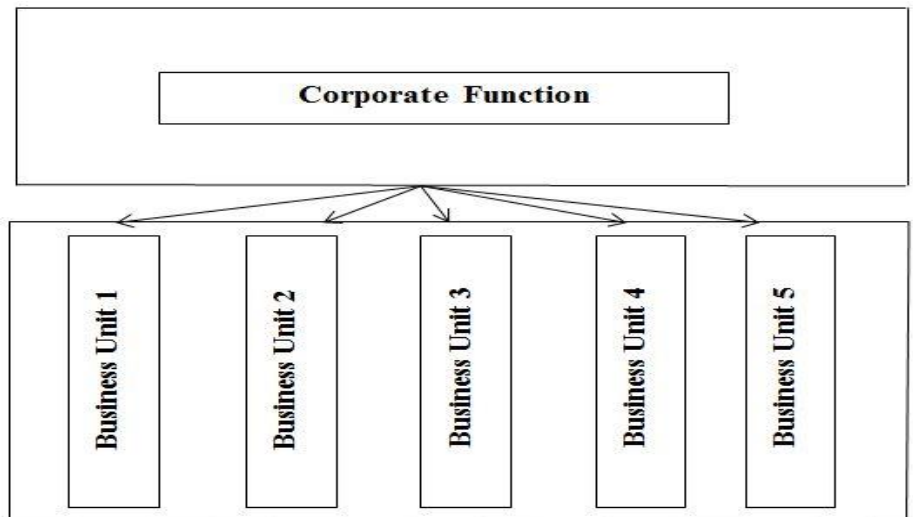
Figure 3: type C (Takaoka 2011)

## 2.3. Types of diversified corporations in terms of kind of diversification

Diversified corporations can be classified into two main types. The first type is about related diversification, namely, A process that takes place when a business expands its activities into product lines that are similar to those it currently offers. For example, a manufacturer of computers might begin making calculators as a form of related diversification of its existing business ("What Is Related Diversification? Definition and Meaning - BusinessDictionary.Com" n.d.). The second type of diversification is about unrelated diversification, namely, a manufacture of diverse products which have no relation to each other. An example of unrelated diversification in a business could be a toy manufacturer that is also manufacturing industrial wiring for the construction industry ("What Is Unrelated Diversification? Definition and Meaning - BusinessDictionary.Com" n.d.). The unrelated diversification is often called "conglomeration". According to Wade and Gravill (2002); related diversification happens when a parent and its subsidiaries operate in congruent area, while unrelated diversification takes place when a parent and its subsidiary operate in dissimilar area(Wade and Gravill 2003). Evidently, a basis on which the relation in a corporation is defined is of paramount importance for discerning types of diversification. As a matter of fact, when, the relation basis changes, the type of diversification changes as well. The relation basis can range from end products to resources. When it is taken as end products, a large number of business units can be supposed as unrelated as represented in figure 4. But, by putting diversification based on resource based relation, all business units can be supposed as related as shown in figure 5.

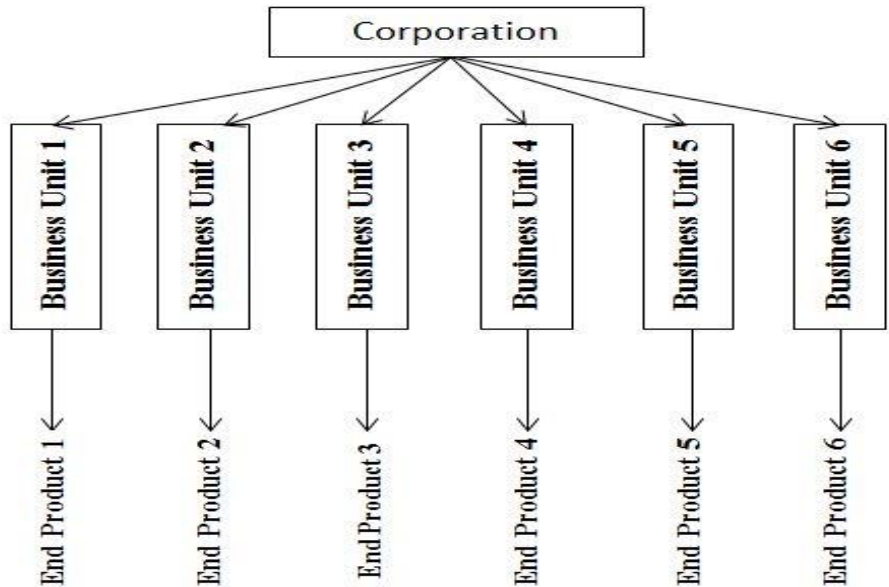

Figure 4: Unrelated diversification based on end product

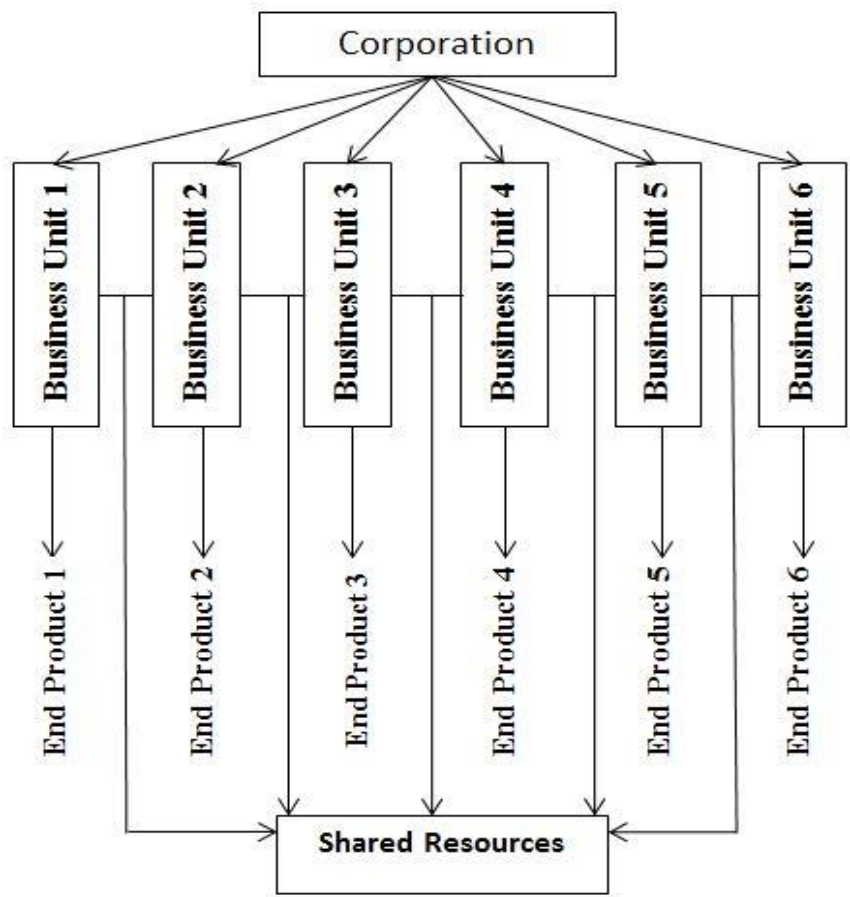

Figure 5: related diversification based on resource relation

.
Regarding figure (5), business units can be linked in terms of shared resources. To understand this kind of diversification more clearly, the concept of resource has to be explicated in details

**2.4 Typology of resources**

Several definitions have been offered regarding the concept of resources. Barney (1991) classifies resources as physical, human and organizational resources. Physical resources include assets, equipment and plants, human resources entail all human related factors such as knowledge base, intellectual capitals, experience and managerial qualifications. Organizational resources include those resources which are organization specific such as organization's culture, image and reputation (Barney 1991). Researchers such as Amit et al. (1993); Collis, (1994) and Helfat et al. (2007) have defined resources as stocks of available factors that are owned, controlled, or accessed on a preferential basis by the firm (Amit and Schoemaker 1993; Collis 1994; Helfat et al. 2009).
 Hall (1992-93) and Katkalo et al. (2010) contend that Resources can be in having (stock) or doing (flow) forms (Hall 1992, 1993; Katkalo, Pitelis, and Teece 2010). "Having" resources are both tangible —like location, material, building, inventory, machinery, and low-skill people—and less tangible—like patents, databases, licenses, brand, and copyright. These resources are normally available to most firms, tradable in the market, and exogenous. In contrast, "doing" resources are intangible — skill-based, firm specific, not tradable in market — and endogenous. Examples of "doing" resources are capabilities incorporated in organizational and managerial processes, like product development, technology development, and marketing. Javidan (1998) defined resources as building blocks of competences and inputs to value chain of organization (Javidan 1998). Noori et al. (2012) have proposed a schematic framework of resource classification of the company. This framework is shown through figure 6.

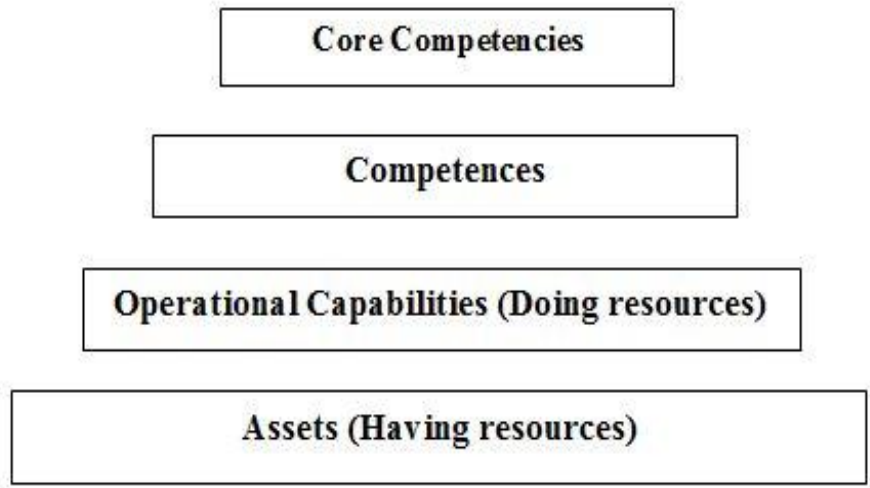

Figure 6: Resource base of the company, a hierarchical classification (Tidd 2012)

A competence is associated with the ability of business units to make coherence among functions and coordinate their capabilities. An example of competence can be the development of a new product through combination of capabilities of management information system, marketing and research and development. According to Prahalad and Hamel (1990)," Core competencies are the collective learning in the organization, especially how to coordinate diverse production skills and integrate multiple streams of technologies"(Prahalad and Hamel 1990). A core competency is a type of competency which possesses fives basic features of (I) Rarity,(II) Value creation, (III) Non substitutability, (IV) highly difficulty for imitation and (V) extendibility (Prahalad and Hamel 1990; Özbağa 2013). Moreover, a generic classification of resources can be presented as figure 7. As can be seen in this figure, resources have been classified into two types of intangible and tangible. Intangible resources share the nature of "doing resources". This type of resources includes materials, stocks, machinery and every other physical embodiment. Another type of resources is related to tangible ones. Tangible resources share the nature of "having resources" which entails intellectual capitals, Knowledge base, employees' experiences, quality of operational processes and other non-physical characteristics.

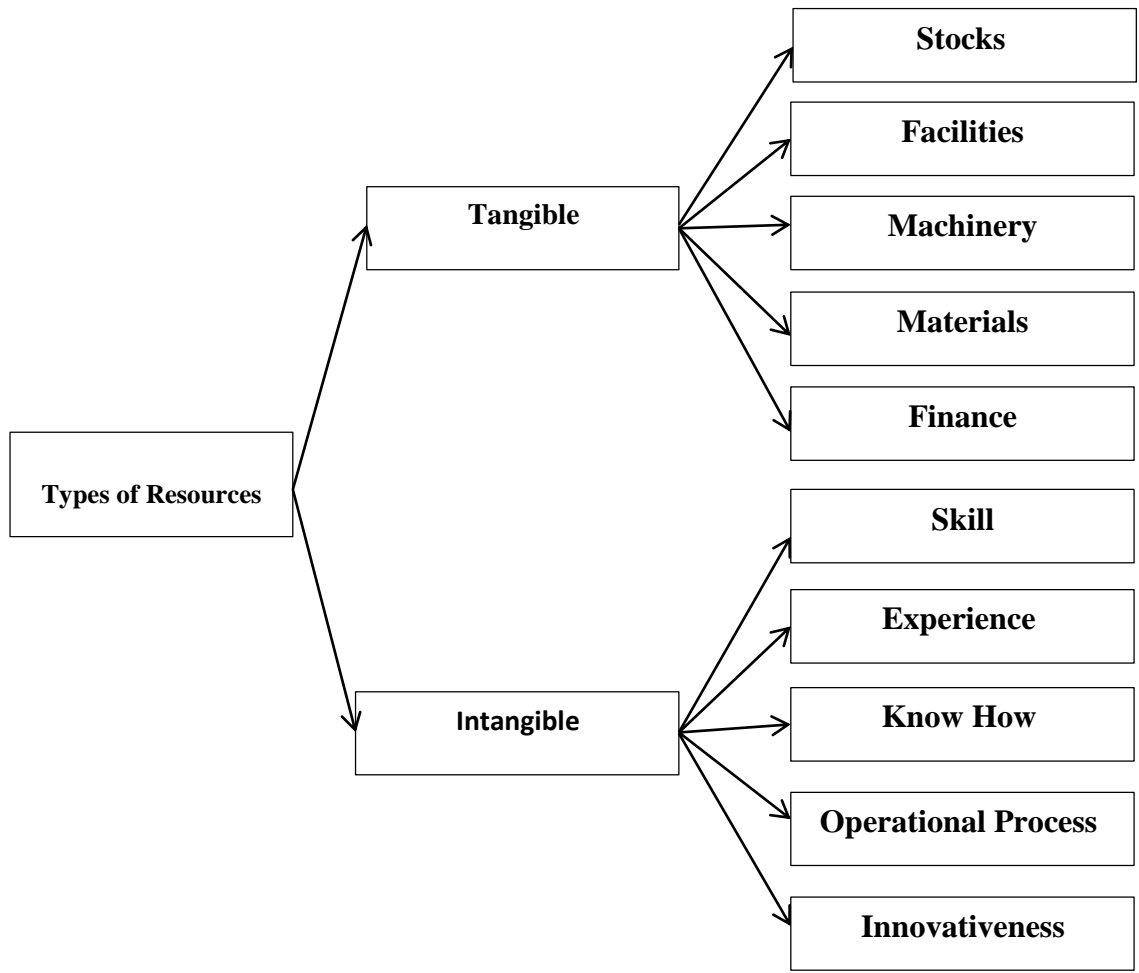

Figure 7: Generic classification of resources

## 2.5. Horizontal strategy

Synergy matters a lot for corporate-lever decision makers of a diversified corporation, to create it among business units, the horizontal strategy has been recognized and recommended as a very effective process. This strategy has been defined by a number of scholars. Porter (1985) holds that horizontal strategy is a way by which a multi-business corporation can combine those differed but related businesses that can create value (Porter 1985). Hax and Najluf (1996) defined it as "**Adding value** beyond the simple sum of independent business contributions"(Hax and Majluf 1996). Kaplan and Norton (2006) believe that horizontal strategy is a way through which a corporation can "add value to its collection of business units and shared service units" (Kaplan and Norton 2006). Shovarini and associates define it as" a set of decisions and actions with the aim of developing and exploiting competitive advantage through sharing tangible and intangible resources among businesses of a multi business firm" (Shovareini, Alborzi, and Mohammad, n.d.).However, the most complete definition of horizontal strategy is believed to belong to Takaoka (2012) where he defines it as "Adding value beyond the simple sum of **independent** but related business units and shared service units by combining their **assets** and **capabilities**" (Takaoka 2011).

As can be inferred from above mentioned definitions, two main concepts of synergy and resource sharing constitute horizontal strategy. Put simply as Takaoka pointed out, "Synergy is the difference between corporate value and the simple sum of the value of business units and shared service units" which can be presented as figure 8.

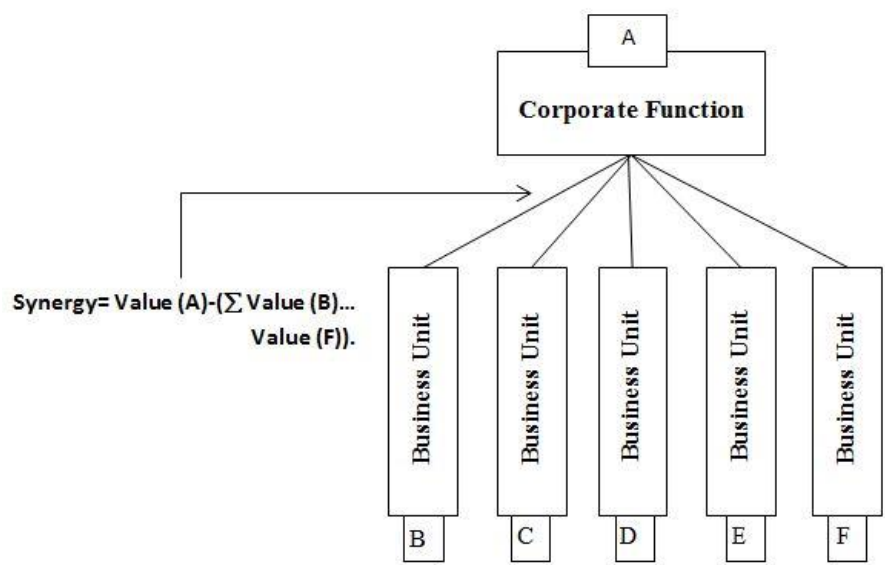

Figure 8: Synergy and horizontal strategy (Takaoka 2011)

When synergy is created across business units, it results in competitive advantage for businesses. Horizontal strategy seeks to create synergy through sharing resources among business units.

## 2.6. Levels of resource sharing

As shown earlier in figure (3), resources can be defined into two types of intangible and tangible in terms of which scholars such as porter (1985) and Ensign (2004) have offered two schemes for resource sharing levels (Porter 1985; Ensign 2004). Porter classifies levels of resource sharing to activity sharing and skill sharing as shown in figure

9. Activity sharing is the easiest part of synergy creation. Shared services units are a good example of activity sharing.

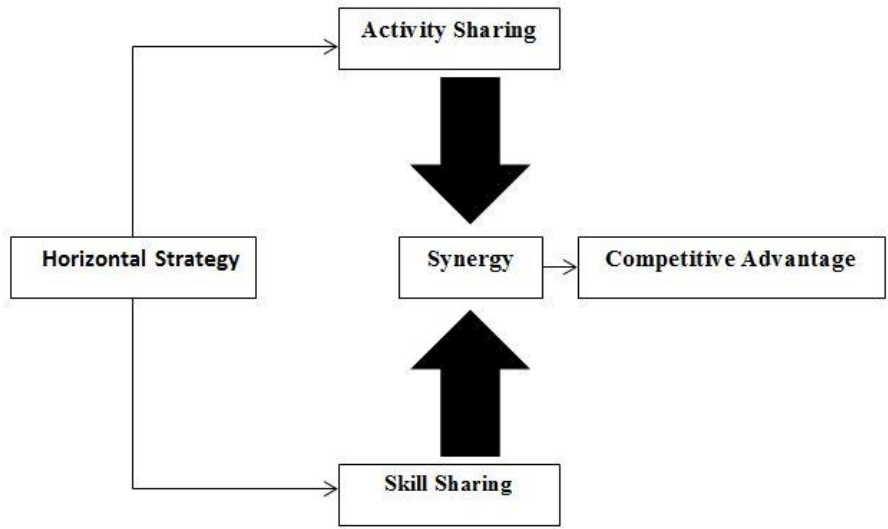

Figure (9) Porter's view of resource sharing (Porter 1985)

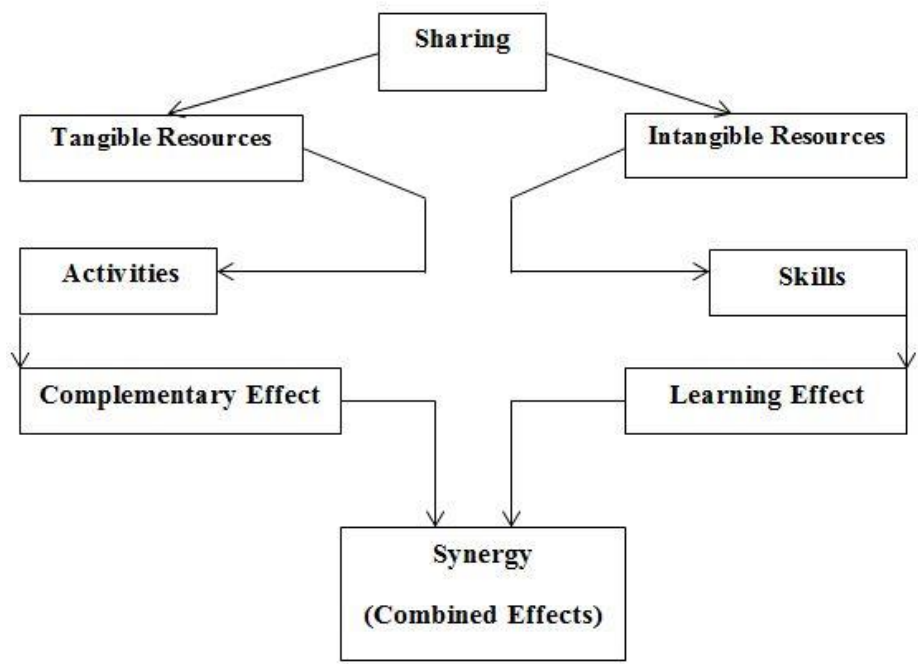

Figure (10) Ensign's view of resource sharing (Ensign 1998)

Skill sharing is associated with sharing of intangible resources among business units. Synergy results from parallel execution of these two types of sharing. It means none of these sharing can per se result in synergy and then competitive advantage. Ensign (1998) has discussed this issue in a relatively similar way. According to him, those related resources whose sharing can lead to synergy are defined as interrelationships and the main purpose of horizontal strategy is to recognize these interrelationships and sharing them (Ensign 1998). Ensign's view of resource sharing is presented in figure 10. As can be seen from it, synergy is a direct result of learning effects generated through intangible resource sharing and complementary effect caused by through tangible resource sharing.

## 3. Success Probability Assessment Index System (SPAIS)

To assess the success vale of horizontal strategy execution, a success probability assessment index system (SPAIS) should be developed. To do so, a two-step procedure has been employed. In the first step, Since the successful execution of horizontal strategy is to a large part dependent on the quality of resource sharing among related business units, the first step deals with identifying influential factors of resource sharing across business units. So, after interviewing some subject matter experts and studying related literature(Teimouri, Emami, and Hamidipour 2011; Zhang, Faerman, and Cresswell 2006; Hendriks 1999; Spender 1996; Jacob and Ebrahimpur 2001), 25 variables related to the success probability of resource sharing across business units were captured and classified to three main groups as represented in table 1.

Table 1: Variables related to successful resource sharing among business units

| Variables | Content | |
|---|---|---|
| X1 | The fit of technology with organizational needs | Through literature review |
| X2 | X2: Updating of machinery & Equipment | |
| X3 | X3: Ease of use of technology | |
| X4 | X4: Strategies' fit with environmental change | |
| X5 | X5: Existence of clear strategies in organizations | |
| X6 | X6: Employee's accurate perception about organization's strategies | |
| X7 | X7: Coherence among organization's strategies | |
| X8 | X8: Existence of informal communication in organization | |
| X9 | X9: Existence of shared values among employees | |
| X10 | X10: Existence of sense of trust & cooperation | |
| X11 | X11: Superiority of organization's interest to individual one | |
| X12 | X12: Existence of shared values among employees | |
| X13 | X13: Existence of short and effective communication channels | |
| X14 | X14: Existence of effective regulations | |
| X15 | X15: Decentralization in decision making | |
| X16 | X16: Existence of a flexible structure | |
| X17 | X17: Coordination among organizational various activities | |
| X18 | X18: Obliteration of redundancies and bottlenecks | |
| X19 | X19: Separation of job duties | |
| X20 | X20: Existence of an evaluation system | |
| X21 | X21: Diversity of job duties | |
| X22 | X22: Flexibility | Through interviewing subject matter experts |
| X23 | X23: Suitable attitude towards employees | |
| X24 | X24: Leadership capability | |
| X25 | X25: Submission to scientific management rules | |

Now, variables included in table 1 are distinctively divided into three groups as presented in table 2.

Table 2: Main groups and their related variables

| Main groups | Related variables |
|---|---|
| A: Organizational Technology &Strategy | X1: The fit of technology with organizational needs<br>X2: Updating of machinery & Equipment<br>X3: Ease of use of technology<br>X4: Strategies' fit with environmental change<br>X5: Existence of clear strategies in organizations<br>X6: Employee's accurate perception about organization's strategies<br>X7: Coherence among organization's strategies |
| B: Organizational Culture & Structure | X8: Existence of informal communication in organization<br>X9: Existence of shared values among employees<br>X10: Existence of sense of trust & cooperation<br>X11: Superiority of organization's interest to individual one<br>X12: Existence of shared values among employees<br>X13: Existence of short and effective communication channels<br>X14: Existence of effective regulations<br>X15: Decentralization in decision making<br>X16: Existence of a flexible structure |
| C: Organizational Process & Management | X17: Coordination among organizational various activities<br>X18: Obliteration of redundancies and bottlenecks<br>X19: Separation of job duties |

|  | X20: Existence of an evaluation system |
|  | X21: Diversity of job duties |
|  | X22: Flexibility |
|  | X23: Suitable attitude towards employees X24: Leadership capability |
|  | X25: Submission to scientific management rules |

In the second step, the information included in table 2 is used for developing success probability assessment index system (SPAIS) of executing horizontal strategy in a multi-business firm. However, though these variables are regarded to be effective on the success of horizontal strategy execution, their effect is not the same. Therefore, some factors which are referred to as critical success factors (CSFs) have higher effects than others and managers should pay very much attention to them.

The concept of CSFs was first proposed by John F Rockart (1979) from MIT Sloan school of management. He defined CSFs as variables and conditions that considerably affect the success of any organizational system (Rockart 1979). Methodologies such interview with experts and decision makers (Zhou, Huang, and Zhang 2011) and Delphi(Shen, Lin, and Tzeng 2011) have been used identify CSFs. Since such methodologies can be easily impacted by subjective biases of the participants, the multi criteria decision making (MCDM) methodology has a better performance for discerning CSFs. As an efficient MCDM methodology, the decision making trial and evaluation laboratory (DEMATEL) has been utilized to construct SVAIS.

### 3.1. DEMATEL

DEMATEL was first proposed by Gabus and Fontela in the early 1970s (Gabus and Fontela 1973, 1972). This method can combine the opinions of subject matter experts into an integrated model showing the causal relationships among the factors of the system under study. Through DEMATEL, it can be understood which factors have a positive net effect on system and which factors don't have such an effect. The factors having a considerable effect are selected as the CSFs (Zhou, Huang, and Zhang 2011). After development of a DEMATEL via the opinions of 10 experts, the final result of DEMATEL to construct a SPAIS (those variables that have a positive net effect) has been presented in table 3.

Table 3: Constructed SPAIS for executing horizontal strategy in a diversified firm

| Main groups | Related variables |
|---|---|
| A: Organizational Technology &Strategy | X1: The fit of technology with organizational needs<br>X3: Ease of use of technology<br>X4: Strategies' fit with environmental change<br>X6: Employee's accurate perception about organization's strategies |
| B: Organizational Culture & Structure | X8: Existence of informal communication in organization<br>X10: Existence of sense of trust & cooperation<br>X12: Existence of shared values among employees<br>X15: Decentralization in decision making |
| C: Organizational Process & Management | X17: Coordination among organizational various activities<br>X19: Separation of job duties<br>X20: Existence of an evaluation system<br>X24: Leadership capability |

## 4. Case study

Since its establishment in the late 1980s, SINACO[1] has grown to be currently known as one of the largest multi-business corporations of Iran. As a diversified corporation, its main purpose is to invest in Iran's various industrial fields. SINACO owns 9 holdings each of which has a number of subsidiaries. SINACO's holdings are presented and described in table 4.

Table 4: SINACO's holdings and number of their respective firms

| Name | Main Focus | # Of Firms |
|---|---|---|
| SINACO1 | Oil, gas, and Petrochemicals | 17 |
| SINACO2 | Pharmaceuticals | 7 |
| SINACO3 | Bank and Insurance | 14 |
| SINACO4 | Cement | 17 |
| SINACO5 | Paper, Wood and Fiber | 12 |
| SINACO6 | ceramic and Tile | 8 |
| SINACO7 | Rail way, shipping | 7 |

---

[1] - The name of company has been changed to protect its anonymity.

| SINACO8 | Energy | 10 |
| SINACO9 | Modern technologies | 5 |
| Sum | | 97 |

For a long time, selection and development of corporate business units have been the focus of attention of several strategists, but since recent world financial crisis, corporate strategists have changed their attention towards the ways corporations achieve competitive advantage on their business levels. As a way for gaining competitive advantage in large diversified corporations, horizontal strategy has been frequently used by several large diversified corporations such as General Electric, Virgin Group, Proctor & Gamble and Kirin Group (Takaoka 2011).

Since horizontal strategy deals with resource sharing among those business units that are related in terms of skill and activity and independent in terms of market, a number of business units subordinating SINACO were selected for this study. Table 5 offers these business units and their related superordinate holding company.

Table 5: Selected Business units and their related superordinate holding company

| Business unit | Superordinate holding |
|---|---|
| Banking | SINACO3 |
| Insurance | SINACO3 |
| Investment | SINACO3 |
| Brokering | SINACO3 |

As can be seen in table 6, SINACO3 is the holding company in which the success probability of executing horizontal strategy is going to be studied. This company holds 14 firms which can be organized into 4 business units. As mentioned earlier, these business units are independent in terms of market though related in terms of resources. As shown in Table 6, for each BU, one of its related firms is chosen.

Table 6: Selected BUs and their related firms

| Firm | BU |
|---|---|
| SINACO3-Bank | Banking |
| SINACO3- Insurance | Insurance |
| SINACO3- Investment Firm | Investment |
| SINACO3- Brokerage | Brokering |

For each firm, 20 employees including director, assistants and experts were selected and using information included in table 3, a survey was developed and distributed among them. Before distributing the survey, it was justified and explained to all of firms' targeted respondents and distributed to them from January 18, 2015 to January 25, 2015. The due time of questionnaire's reception was set for 10 days later (i.e. February 5, 2015). Among all 280 persons that received the survey just 0.75 of them responded to it up to the end of due time.

## 5. Model Development

### 5.1 Artificial Neural Networks (ANN)

In the field of artificial intelligence (AI), an artificial neural network (ANN) is known as a powerful computational data model that is able to capture and represent complex input/output relationships. The motivation for the development of neural network technology stemmed from the desire to develop an artificial system that could perform "intelligent" tasks similar to those performed by the human brain ("Neural Network Software, Data Mining, Neural Networks, Distributed Computing, Artificial Intelligence, NeuroSolutions, TradingSolutions, Trader68, OptiGen Library, Neural Network Course | NeuroDimension, Inc." n.d.). ANNs are basically presented as systems of interconnected "neurons" that are able to compute values from inputs, and have the capability of machine learning as well as pattern recognition because of their adaptive nature.

In real world problems, ANNs have been applied in a wide range of fields ranging from aerospace engineering to banking industry. Hakimpour and associates has conducted a research on ANNs' applications in management in which they have classified its applications based on four main areas and their related problem types. Table 7 shows this classification (Hakimpoor et al. 2011)

Table 7: ANNs' reported applications

| Main Area I | Marketing and Sales |
|---|---|
| **Problem Type** | Forecasting costumer response, Market development forecasting, Sales forecasting, Price elasticity modeling, Target marketing, Customer satisfaction assessment, |

| | Customer loyalty and retention , Market segmentation, Customer behavior analysis, Brand analysis, Market basket analysis, Storage layout, Customer gender analysis, Market orientation and performance, Marketing strategies, strategic planning and performance, Marketing data mining, Marketing margin estimation, Consumer choice prediction, Market share forecasting. |
|---|---|
| Main Area II | Finance and Accounting |
| **Problem Type** | Financial health prediction, Compensation assessment, Bankruptcy classification, Analytical review process, Credit scoring, Signature verification, Risk assessment, Forecasting, Stock trend classification, Bond rating, Interest rate structure analysis, Mutual found selection, Credit and evaluation. |
| Main Area III | Manufacturing and Production |
| **Problem Type** | Engineering design, Quality control, Storage design, Inventory control, Supply chain management, Demand forecasting, Monitoring and diagnosis, Process selection. |
| Main Area IIII | Strategic Management and Business Policy |
| **Problem Type** | Strategic planning and performance, Assessing decision making, Evaluating strategies |

The other examples of ANN applications can be seen in (Paliwal and Kumar 2009; Li 1994; Meireles, Almeida, and Simões 2003; Murray 1995; Widrow, Rumelhart, and Lehr 1994).

## 5.2. Developed ANN

The ANN developed in this paper is represented in Figure 11. All input vectors of proposed ANN have 12 elements of SVAIS. The number of these input vectors is equal to that of respondents (i.e. 75). The proposed ANN has 10 neurons (i.e., nodes) in its hidden layers each of which has a hyperbolic tangent sigmoid transfer function. This function's structure is shown as:

$$f(n) = \frac{e^n - e^{-n}}{e^n + e^{-n}} \tag{1}$$

Mathematically, for an interval of [-10, 10], it compresses all of its inputs to a range from -1 to +1, as it is shown in figure 12.

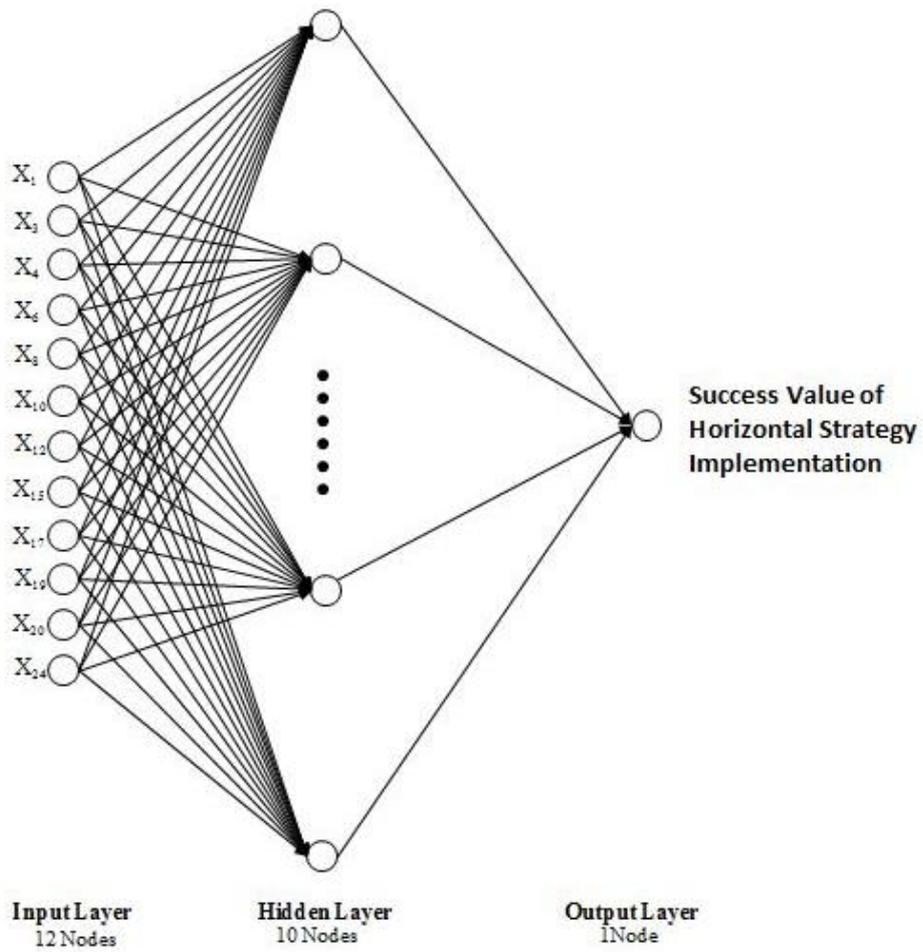
Figure 11: Proposed ANN

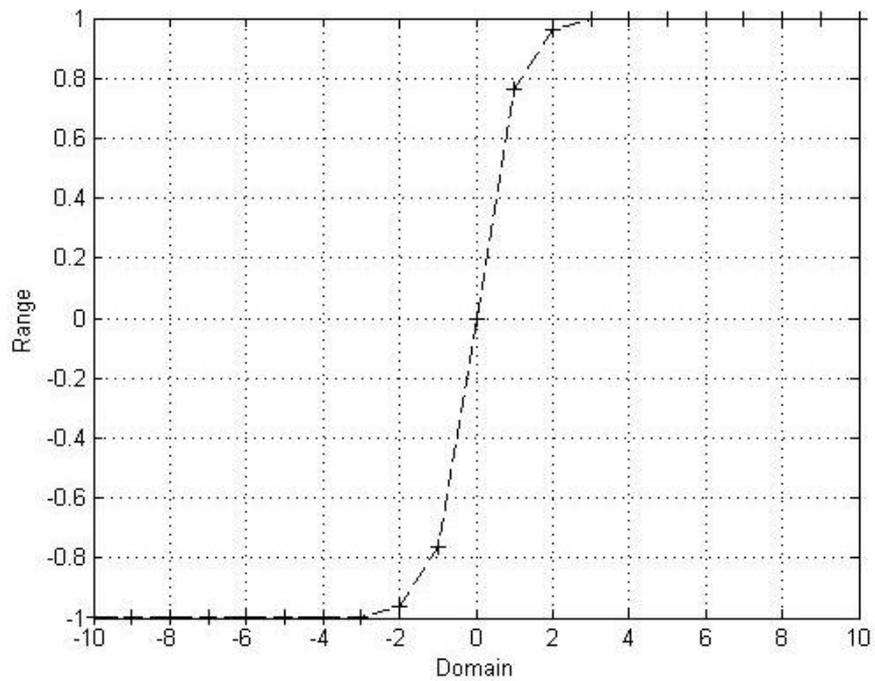
Figure 12: Hidden layer's function diagram

In the output layer there is just 1 neuron equal to the success probability of horizontal strategy execution. The transfer function used in this neuron is

$$f(n) = n \qquad (2)$$

This function gives back whatever it takes. The diagram of this function is show in Figure 13.

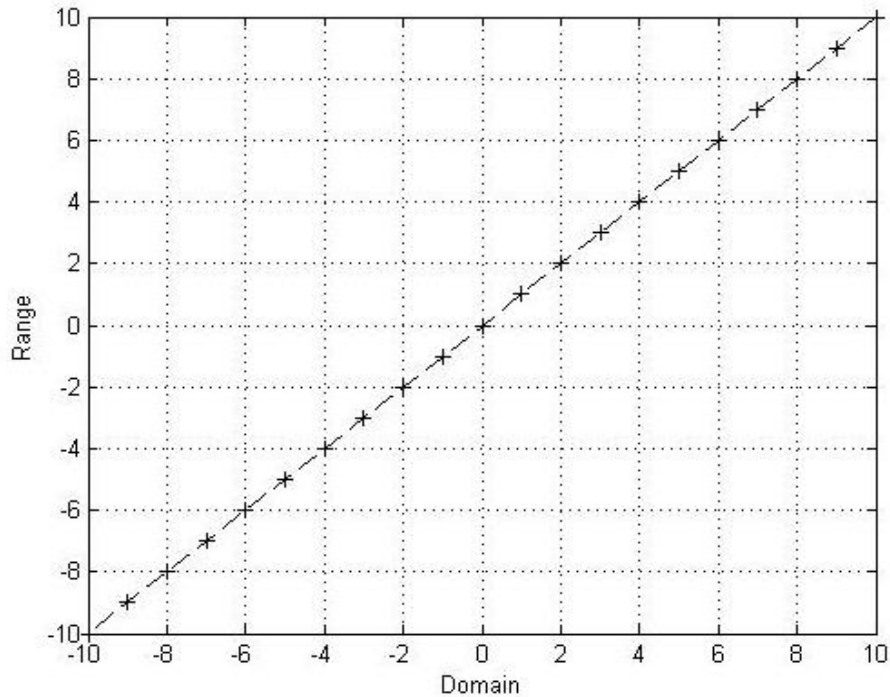

Figure 13: Output layer's function diagram

The performance function of proposed model is shown as

$$MSE = \frac{\sum_{i=1}^{N} e_i^2}{N} \qquad (3)$$

Where $e_i$ stands for the error of $i_{the}$ neuron and N represents the number of all neuron of the network. Actually, the model's most important purpose is to reduce this performance function as much as possible. To do so, Levenberg-Marquardt back propagation algorithm has been found very efficient. This algorithm enables the network to update its parameters (i.e. weights and biases) in order to reduce performance function value. Parameter updating in turn is done through an iteratively training manner.

Specifically, this mechanism enables the network to determine the gradient of performance function and by means of its training function updates its parameters for reducing performance function. According to(Hagan, Demuth, and Beale 1996) , the back propagation algorithm is explicated as following:

**5.2.1. Model's Back Propagation Algorithm**

Clearly, in a multilayer network the output of one layer becomes an input for following one. This operation can be described by

$$a^{m+1} = f^{m+1}(W^{m+1}a^m + b^{m+1}) \quad m = 0,1,...,M-1 \qquad (4)$$

Where M represents network's number of layers, first layer's neurons get external inputs

$$a^0 = p \qquad (5)$$

by which the starting point is provided for equation (4). The outputs of the last layer's neurons are considered as network's outputs:

$$a = a^M \qquad (6)$$

Mean Squared Error (MSE) as the performance function is used by back propagation. A set of examples of network's proper behavior are provided for it:

$$\{p_1,t_1\},\{p_2,p_1\},......,\{p_Q,t_Q\} \qquad (7)$$

Where $P_q$ represents a network input and the corresponding target output is indicated by $t_q$. When each input enters the network, the network's output is compared with its target and in this case, algorithm has to adjust the parameters of network to minimize the value of MSE.

$$F(x) = E[e^2] = E[(t-a)^2] \qquad (8)$$

Where x represents the vector of weights and biases of network and if the network has a number of outputs, this can be generalized to

$$F(x) = E[e^T e] = E[(t-a)^T (t-a)] \tag{9}$$

Then, it will approximate the MSE by

$$\hat{F}(x) = (t(k) - a(k))^T (t(k) - a(k)) = e^T(k) e(k) \tag{10}$$

Where the squared error at iteration k has replaced the expectation of the squared error. For calculating the steepest algorithm that can be used for calculating the approximate MSE is:

$$w_{i,j}^m(k+1) = w_{i,j}^m(k) - \alpha \frac{\partial \hat{F}}{\partial w_{i,j}^m} \tag{11}$$

$$b_i^m(k+1) = b_i^m(k) - \alpha \frac{\partial \hat{F}}{\partial b_i^m} \tag{12}$$

Where the $\alpha$ represents the learning rate.

Now the partial derivatives of Eq. (11) and Eq. (12) should be computed by chain rule method.

$$\frac{\partial \hat{F}}{\partial w_{i,j}^m} = \frac{\partial \hat{F}}{\partial n_i^m} \times \frac{\partial n_i^m}{\partial w_{i,j}^m} \tag{13}$$

$$\frac{\partial \hat{F}}{\partial b_i^m} = \frac{\partial \hat{F}}{\partial n_i^m} \times \frac{\partial n_i^m}{\partial b_i^m} \tag{14}$$

The second term in each of these equations can be easily computed, since the net input to layer m is an explicit function of the weights and bias in the layer:

$$n_i^m = \sum_{j=1}^{s^{m-1}} w_{i,j}^m a_j^{m-1} + b_i^m \tag{15}$$

Therefore

$$\frac{\partial n_i^m}{\partial w_{i,j}^m} = a_j^{m-1}, \frac{\partial n_i^m}{\partial b_i^m} = 1 \tag{16}$$

If we now define the sensitivity of $\hat{F}$ to changes in the ith element of the net input at layer m

$$s_i^m = \frac{\partial \hat{F}}{\partial n_i^m} \tag{17}$$

Now Eq. (13) and Eq. (14) can be simplified to

$$\frac{\partial \hat{F}}{\partial w_{i,j}^m} = s_i^m a_j^{m-1} \tag{18}$$

$$\frac{\partial \hat{F}}{\partial b_i^m} = s_i^m \tag{19}$$

We can now express the approximate steepest descent algorithm as

$$w_{i,j}^m(k+1) = w_{i,j}^m(k) - \alpha s_i^m a_j^{m-1} \tag{20}$$

$$b_i^m(k+1) = b_i^m(k) - \alpha s_i^m \tag{21}$$

In matrix form this becomes

$$w^m(k+1) = w^m(k) - \alpha s^m (a^{m-1})^T \tag{22}$$

$$b^m(k+1) = b^m(k) - \alpha s^m \tag{23}$$

Where

$$s^m = \frac{\partial \hat{F}}{\partial n^m} = \begin{bmatrix} \dfrac{\partial \hat{F}}{\partial n_1^m} \\ \dfrac{\partial \hat{F}}{\partial n_2^m} \\ \cdot \\ \cdot \\ \dfrac{\partial \hat{F}}{\partial n_m^m} \end{bmatrix} \quad (24)$$

Now, Sensitivities should be computed. To compute $s^m$, the rule chain method has to be used again. It is this process that gives us the term back propagation, because it describes a recurrence relationship in which the sensitivity at layer m is computed from the sensitivity at layer m+1.

To derive the recurrence relationship for the sensitivities, the following Jacobian Matrix is used:

$$\frac{\partial n^{m+1}}{\partial n^m} = \begin{bmatrix} \dfrac{\partial n_1^{m+1}}{\partial n_1^m} & \dfrac{\partial n_1^{m+1}}{\partial n_2^m} & \cdots & \dfrac{\partial n_1^{m+1}}{\partial n_{s^m}^m} \\ \dfrac{\partial n_2^{m+1}}{\partial n_2^m} & \dfrac{\partial n_2^{m+1}}{\partial n_2^m} & \cdots & \dfrac{\partial n_2^{m+1}}{\partial n_{s^m}^m} \\ \cdot & \cdot & & \cdot \\ \cdot & \cdot & & \cdot \\ \cdot & \cdot & & \cdot \\ \dfrac{\partial n_{s^{m+1}}^{m+1}}{\partial n_2^m} & \dfrac{\partial n_{s^{m+1}}^{m+1}}{\partial n_2^m} & \cdots & \dfrac{\partial n_{s^{m+1}}^{m+1}}{\partial n_{s^m}^m} \end{bmatrix} \quad (25)$$

By considering the i, j element of the above matrix, its expression will be as following:

$$\frac{\partial n_i^{m+1}}{\partial n_j^m} = \frac{\partial \left( \sum_{t=1}^{s^m} w_{i,t}^{m+1} a_t^m + b_i^{m+1} \right)}{\partial n_j^m} = w_{i,j}^{m+1} \frac{\partial a_j^m}{\partial n_j^m} = w_{i,j}^{m+1} \frac{\partial f^m(n_j^m)}{\partial n_j^m} = w_{i,j}^{m+1} \dot{f}^m(n_j^m) \quad (26)$$

Where

So the Jacobian matrix should be written as

$$\dot{f}^m(n_j^m) = \frac{\partial f^m(n_j^m)}{\partial n_j^m} \quad (27)$$

$$\frac{\partial n^{m+1}}{\partial n^m} = W^{m+1} \dot{F}^m(n^m) \quad (28)$$

When

$$\dot{F}^m(n^m) = \begin{bmatrix} \dot{f}^m(n_1^m) & 0 & 0 \\ 0 & \dot{f}^m(n_2^m) & 0 \\ 0 & 0 & \dot{f}^m(n_{s^m}^m) \end{bmatrix} \quad (29)$$

By using chain rule in matrix form, it is possible to write out the recurrence relation for sensitivity as following:

$$s^m = \frac{\partial \hat{F}}{\partial n^m} = \left( \frac{\partial n^{m+1}}{\partial n^m} \right)^T \frac{\partial \hat{F}}{\partial n^{m+1}} = \dot{F}^m(n^m)(W^{m+1})^T \frac{\partial F}{\partial n^{m+1}}$$
$$= \dot{F}^m(n^m)(W^{m+1})^T s^{m+1} \quad (30)$$

To make back propagation completed, the starting point $s^m$ for the recurrence relation f Eq. (30) is needed. This is obtained at the final layer:

$$s_i^M = \frac{\partial \hat{F}}{\partial n_i^M} = \frac{\partial (t-a)^T (t-a)}{\partial n_i^M} = \frac{\partial \sum_{j=1}^{s^M} (t_j - a_j)^2}{\partial n_i^M} = -2(t_i - a_i) \frac{\partial a_i}{\partial n_i^M} \quad (31)$$

Now, Since

$$\frac{\partial a_i}{\partial n_i^M} = \frac{\partial a_i^M}{\partial n_i^M} = \frac{\partial f^M(n_i^M)}{\partial n_i^M} = f^M(n_i^M) \tag{32}$$

It can be written as

$$s_i^M = -2(t_i - a_i) f^M(n_i^M) \tag{33}$$

Which its matrix form is expressed as

$$s^M = -2F^M(n^M)(t-a) \tag{34}$$

Most often, BPs use a gradient descent algorithm for adjusting network's parameters. However, when the dimensions of ANNs get larger and more complicated, the Levenberg-Marquardt algorithm is strongly recommended especially because of its operation speed and accuracy. So, in this paper, a Levenberg-Marquardt back propagation has been used for network training.

## 6. Results Analysis

After writing and solving the proposed model by MATLAB Software, a set of various results was achieved all of which are presented as following:

### 6.1. Types of network performance

Basically there are four types of performance based on which an ANN can be analyzed. These are validation performance, test performance, train performance and network's overall performance. The more these performance values are closer to zero, the better ANN can do forecasting.

#### 6.1.1. Validation performance results

Figure 14 shows the values of test, train and validation performances. Here, the number of epochs is equal to that of times which ANN has been allowed to be trained. As can be seen, all performances have had a descending order up to the second epoch where training performance has had an increase.

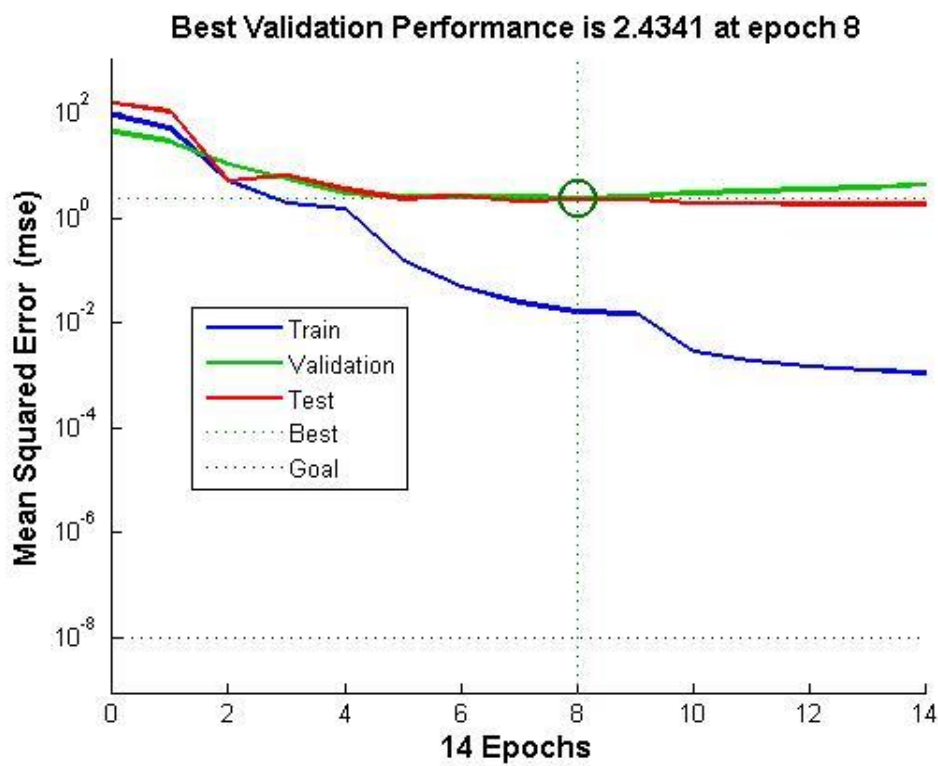

Figure 14: Network performances

Validation performance is the most important indicator for analyzing the network behavior. Actually, when this performance value goes up, it means that the network has started being over trained so its behavior will become unstable or chaotic over time. Therefore, the less validation performance value is, the more stable network's behavior is expected. However, while starting being over trained, the network training operation is stopped where the validation performance has had the least MSE value. As shown in figure 14, the eighth epoch is where the network training operation has been stopped because after this point, the network has reached maximum level of allowed failures (i.e. 6 failures). This is shown in figure 15. However, the best performance value of this model is 2.4341 showing that the network behavior is stable and its generalizability is high.

**6.1.2. Training performance results**

As an indicator for network training quality, in the eighth epoch the value of training performance is 0.0164 showing that the network's training quality is very good in a way that its performance has become better in each of subsequent epochs (figure 14).

**6.1.3. Test performance results**

The test performance which indicates network's learning quality is 2.2430 in the eighth epoch meaning that ANN has a relatively good performance index in this aspect. However, this performance value is acceptable and proves ANN's good learning quality (figure 14).

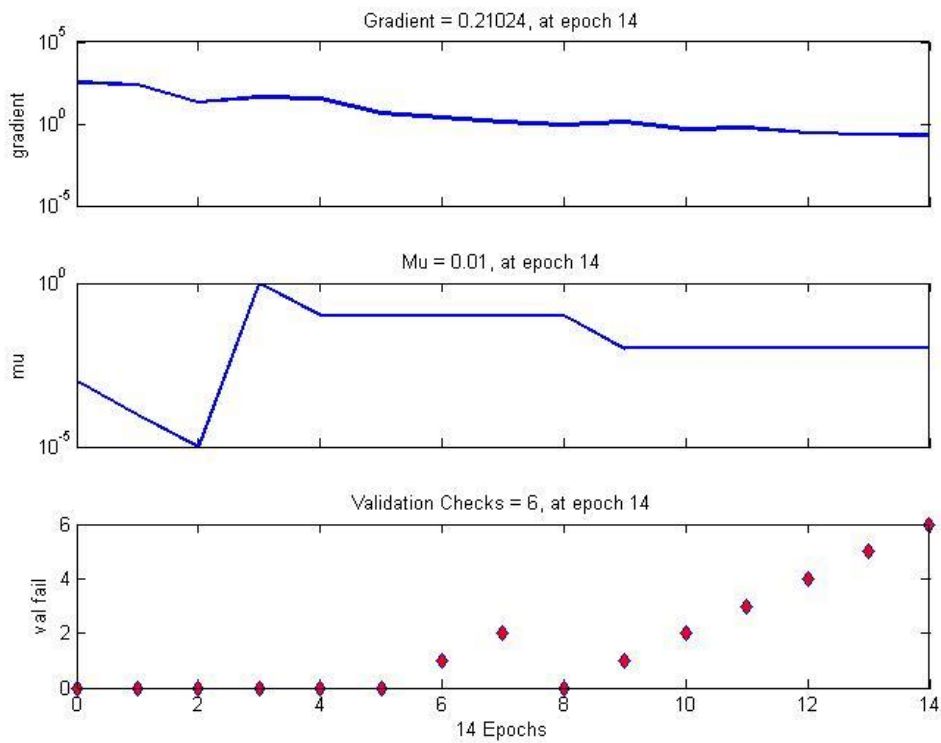

Figure 15: Validation failures

**6.1.4. Network overall performance**

As shown in figure 16, the network's overall performance (MSE) is 0.0.82651 which shows a really good result. The regression value (R=0.98316) indicates a very high correlation between network's outputs and real ones and error histogram parameters such as mean (0.1043) and standard deviation (0.90921) signify that network's total performance relatively follows a normal distribution. However, these results verify the ability of developed model to forecast the success probability of executing horizontal strategy in a diversified corporation with an accuracy of 0.98316.

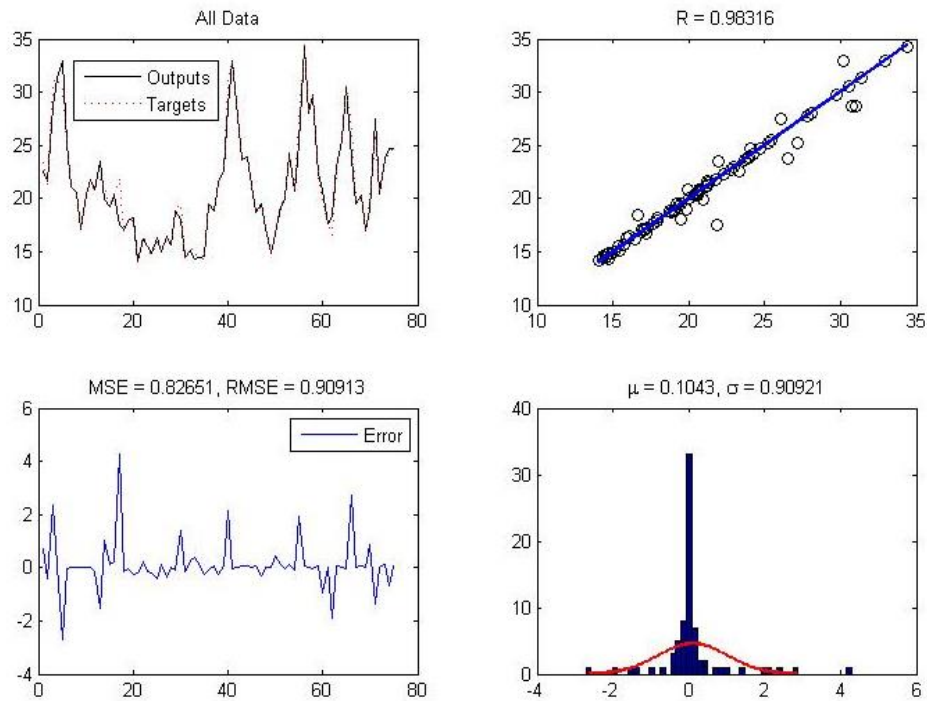

Figure 16: Network's total performance

## 6.2. Error Histogram

Error Histogram of an ANN provides much precious information about its errors. Error Value (EV) and Error Frequency (EF) are two main data that can be extracted from error histogram. The variance of errors also shows that errors can be classified to big and small one in terms of EV. The negative sign of an error for each performance index happens when its outputs are larger than its targets. As shown in figure 17, in training data set which entails 65% of all samples, most of errors are closed to zero (small errors) while the most of errors in validation data set (which includes 20% of all samples) are far from zero (big errors). Errors of test data set which entails 15% of all input samples are more inclined to than validation test. However, network's error distribution is relatively normal with a mean of 0.1043 of and standard deviation of 0.90921.

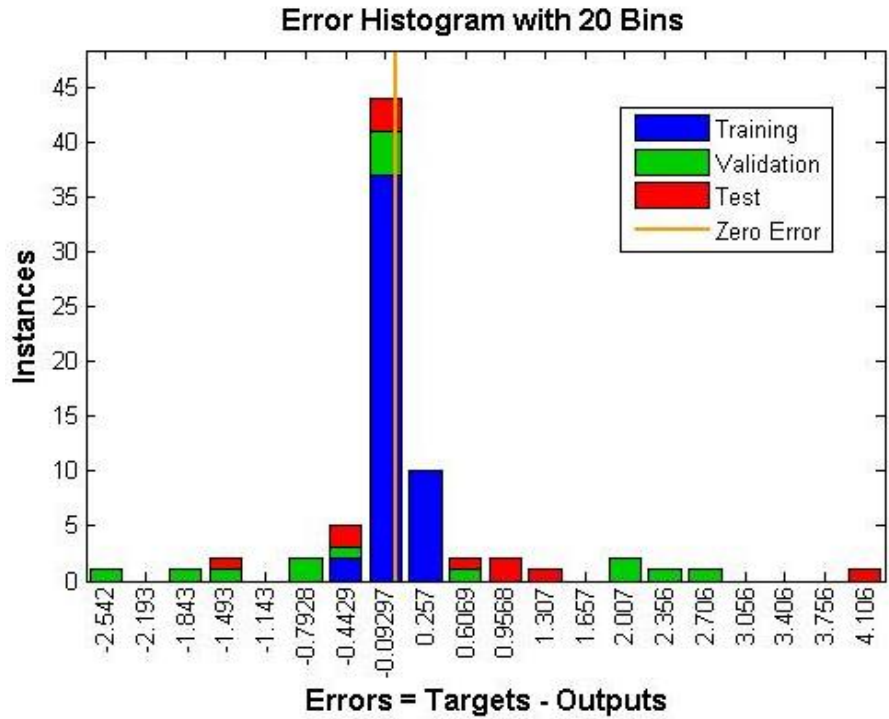

Figure 16: Error histogram

## 7. Conclusion

As a very powerful way for synergy creation at business level, horizontal strategy completely engages business units of a diversified corporation by resource sharing among them. So, executing it without enough information will result in undesirable outcomes. such as failure and financial resources wastage. To manage its execution more confidently, managers should have reliable information about its success probability in advance. The Model developed in this paper is aimed at helping managers to have such precious information and enabling them to forecast the success probability of executing horizontal strategy in a multi-business corporation far better than other classical models. According to a success probability assessment index system (SPAIS) that has been extracted from valid resources and constructed by the decision making trial and evaluation laboratory (DEMATEL) method, an ANN has been designed and trained by a Levenberg-Marquardt back propagation algorithm for enabling corporate managers to forecast the success probability of horizontal strategy execution before starting it at business level. The heighted level of model's accuracy and reliability makes it a very reliable mechanism for measuring the success probability of horizontal strategy execution in advance.

However, the proposed model can be improved in three aspects. The first aspect points to this fact that since ANNs generate complex surfaces of error with several local optima points, even for a simple function approximation problem, gradient research algorithms mainly intend to get trapped in local solutions which are not global. So, back propagation as a gradient research method seems not to be able to offer the best and fastest mechanisms for neural networks training. To prevent ANN from being trapped in local solutions, especially when a nonlinear problem is supposed to be modeled, Meta heuristics such as genetic algorithm (GA) and ant colony optimization (ACO) have been found very effective for ANN training and can be used for this purpose. The second aspect is related to optimal number of ANN's hidden layers' neurons. In other words, the performance of an ANN really improves when its hidden layer's neuron number is optimal. So, there is a strong need for developing a mechanism by which researchers can improve model's architecture. Like first aspect, meta heuristics can also be used for this purpose. The third aspect is about the nature of model's variables which all can be dealt with in a fuzzy manner; therefore, development of a fuzzy ANN is strongly needed. However, pursuing each of these three aspects is of paramount value and can be a subject for future researches.